%% file: main.tex
\documentclass[conference]{IEEEtran}
\usepackage{amsmath,amssymb,amsfonts}
\usepackage{algorithmic}
\usepackage{graphicx}
\usepackage{textcomp}
\usepackage{xcolor}
\usepackage{booktabs}
\usepackage{array}
\usepackage{url}
\usepackage{placeins}
\usepackage{float}

\graphicspath{{figures/}}

\usepackage{tikz}
\usetikzlibrary{positioning,shapes,arrows.meta}

\usepackage[colorlinks=true, citecolor=blue, linkcolor=blue, urlcolor=blue]{hyperref}

\def\BibTeX{{\rm B\kern-.05em{\sc i\kern-.025em b}\kern-.08em
    T\kern-.1667em\lower.7ex\hbox{E}\kern-.125emX}}

\begin{document}

\title{\LARGE \bf
CADENCE: Predicting Realized MAPF Execution Time Beyond Sum of Costs}

\author{
\IEEEauthorblockN{Abhishek Seetharamu\textsuperscript{*}}
\IEEEauthorblockA{\textit{BuildMachineLabs}\\
\href{mailto:abhishekss6363@gmail.com}{mail}}
\and
\IEEEauthorblockN{Badrikanath Praharaj\textsuperscript{*}}
\IEEEauthorblockA{\textit{BuildMachineLabs}\\
\href{mailto:badrikanath.praharaj@gmail.com}{mail}}
\and
\IEEEauthorblockN{Sreeram M.V.\textsuperscript{*}}
\IEEEauthorblockA{\textit{BuildMachineLabs}\\
\href{mailto:mvsreeramblr@gmail.com}{mail}}
\and
\IEEEauthorblockN{Mohan Prakash V}
\IEEEauthorblockA{\textit{IISc, Bengaluru}\\
\href{mailto:vmohanprakashr@gmail.com}{mail}}
\thanks{\textsuperscript{*}These authors contributed equally to this work.}
}

\maketitle
\thispagestyle{empty}
\pagestyle{empty}
\raggedbottom

\begin{abstract}
Multi-Agent Path Finding (MAPF) algorithms are increasingly used to plan motion for robot teams in industrial warehouses and robotic shared workspaces, but standard MAPF algorithm evaluation metrics, such as Sum of Costs (SoC), makespan, and planner runtime, can obscure how planner choices translate into realistic execution performance. We present CADENCE (Coordination- and Action-Driven Estimation for Networked Continuous Execution), a hardware study of this evaluation gap on a fixed $7\times7$ workcell with seven differential-drive robots, asking which features available before execution can best predict final wall-clock completion time. We compare SoC, total planned travel cost, primitive motion burden (how much basic motion the plan requires, such as makespan, turns, consecutive moves, and start--stop transitions), and interaction-aware coordination structure (how much inter-robot coordination the plan induces, such as dependency links, interacting robot pairs, dependency depth, and crowding exposure). To test this, we generate 120 plans across 15 scenarios - 5 Empty, 5 Medium-Random, and 5 Bottleneck---and execute each plan four times, yielding a 480-trial hardware corpus. Using both a scenario-held-out ridge model and a trial-level mixed-effects model, we find that SoC alone is informative but incomplete, while primitive motion burden gives the strongest improvement, reducing held-out error by about 48.6\%--59.8\% in MAE and 44.2\%--61.4\% in RMSE relative to SoC-only models. Interaction-aware coordination features add smaller, less uniform gains, most clearly in the mixed-effects analysis. Across both models and uncertainty checks, primitive motion burden is the most reliable additional signal beyond SoC, suggesting that much of the execution time gap is already visible in the offline plan before any robot starts moving.
\end{abstract}


\section{Introduction}

\begin{figure}[!t]
\centering
\includegraphics[width=\columnwidth]{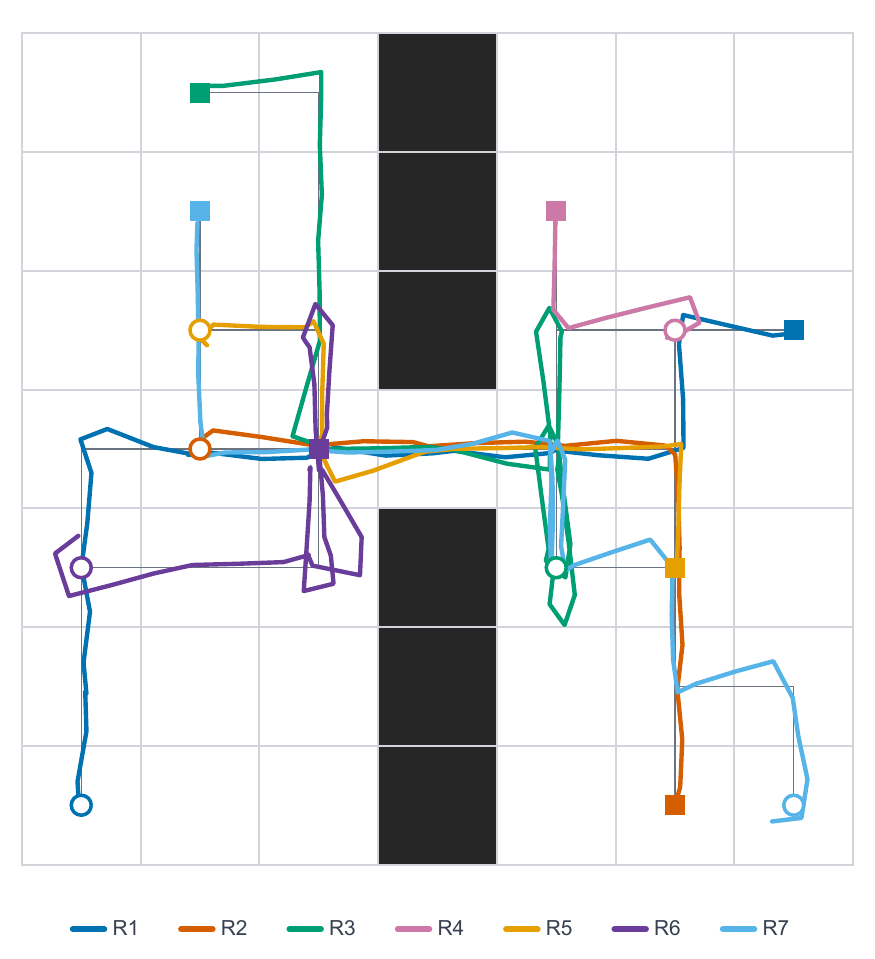}
\vspace{-0.75em}
\setlength{\abovecaptionskip}{0pt}
\setlength{\belowcaptionskip}{0pt}
\caption{\textbf{Hardware OptiTrack trace.}
Color-coded trajectories for Robots 1--7 on a representative $7\times7$
bottleneck execution. Circles mark starts; squares mark goals.}
\label{fig:hardware-trial}
\end{figure}

MAPF algorithms are deployed in warehouse-style multi-robot systems, where many robots navigate shared aisles to reach goals without collisions. In these environments, collision avoidance and efficient goal attainment are critical. A plan may seem efficient during offline computation, but its execution on physical hardware can be slower due to factors beyond the nominal plan cost, such as motion dynamics and coordination during deployment. Effective deployment requires considering both planned costs and the dynamics of motion and coordination among robots. This motivates the central research question: Once an MAPF plan is executed on physical robots, to what extent can the execution time be predicted before deployment? The study examines how much of the observed execution time is evident in the pre-execution plan.

This question matters because physical execution introduces structure that scalar plan costs suppress. Robots must preserve the same-agent action order, satisfy cross-robot precedence constraints, pass through locally contested spaces, and often incur stop--go delays under an execution monitor~\cite{Ma2017, Hoenig2019, Su2024}. As a result, plans with similar SoC can produce materially different wall-clock completion times on hardware. Figure~\ref{fig:hardware-trial} shows one representative bottleneck execution in which the realized trajectories reflect these effects. This motivates the layered view used in this paper: SoC captures a coarse plan-cost trend, primitive motion burden captures how much basic motion a plan demands, and interaction-aware coordination structure captures how much inter-robot coordination load the plan induces.

We study this as a hardware transfer question using Pololu 3Pi+ validated using Optitrack Motion Capture system Prime 13W.  The same agent action order requires each robot to execute its planned actions sequentially; cross-robot precedence constraints may require one robot to wait until another completes a dependent move. Local contention occurs when multiple robots compete for access to the same narrow region or shared space. Repeated waiting behavior can lead to cumulative stop-and-go delays during execution. Figure~\ref{fig:overview} illustrates the hardware platform and the three scenario families used to represent these interaction regimes. Subsequently, a four-model ladder is evaluated using scenario-level held-out validation to determine which plan side descriptors, computable prior to robot movement, best predict the realized execution time on this fixed hardware stack.

\FloatBarrier

\input{figures/fig_study_overview}
\FloatBarrier
\vspace*{0.9em}

This paper makes three contributions:
\begin{itemize}
  \item A real hardware execution time corpus for offline MAPF plans on a fixed multi-robot platform, built with an interaction-regime-stratified scenario library. 
  \item A disciplined three-layer decomposition of plan-side predictors of realized execution time. We separate plan descriptors into three analytically distinct layers: (i) SoC, the standard MAPF plan quality scalar; (ii) primitive motion burden total turn count, consecutive-move runs, and start–stop transitions; and (iii) interaction-aware plan structure   cross-robot precedence count, precedence depth, and crowding exposure
  \item A cross-scenario test of where the execution-time signal lives beyond SoC. In scenarios withheld entirely from training, Primitive motion burden cuts held-out MAE from 2.76 s to 1.42 s beyond SoC; interaction-aware plan structure contributes a smaller, consistent additional predictive signal beyond primitive burden
 .
\end{itemize}

\section{Related Work and Positioning}\label{sec:related}

MAPF algorithm evaluation is often plan-centric: benchmarks and surveys typically 
compare solvers. These comparisons often utilize metrics such as SoC, 
planned makespan, planner runtime, and success rate. Bounded-sub-optimality is also 
assessed in standardized instances~\cite{Stern2019}. These metrics are appropriate 
when the objective is to analyze search algorithms and planner trade-offs. However, 
the present inquiry differs: given a plan intended for execution on a physical robotic 
system, which properties of that plan account for realized execution time beyond its 
nominal cost?
optimal solvers, such as Conflict-Based Search (CBS)~\cite{Sharon2015CBS}, generate plans
with well-defined cost properties. However, cost minimality does not necessarily result
in minimal realized execution time when factors such as robot motion, precedence constraints, 
and cross-robot interaction structure are taken into account~\cite{Wurman2008,Ma2017}.

The closest conceptual precedent is Yan et al.~\cite{Yan2025Tradeoffs}, who
use SMART realistic simulation to study how MAPF planner-design choices
translate into execution performance. Their results show that  SoC
captures the trend of average execution time, that execution related structure 
improves execution time estimates, and that model accuracy can matter more than
aggressive refinement of a simplified objective. We ask the corresponding
hardware question: on a physical robot stack, do pre-execution plan descriptors
beyond SoC improve held-out prediction of realized wall-clock completion time?
The present work, therefore, differs in three ways. First, it measures physical
hardware rather than simulated execution; a diagnosis that holds in simulation
need not survive motor dynamics, tracking noise, and real contention. Second,
it decomposes the missing signal into primitive motion burden and
interaction-aware coordination structure. Third, it tests a compact feature 
ladder under scenario-held-out validation, so the claim is a platform-scoped 
hardware transfer result 

More broadly, this study relates to realistic execution and dependency-graph
methodologies for MAPF~\cite{Ma2017,Hoenig2019,Varambally2022,Chen2024,Su2024,
Duhan2026P3GASUS}. These works show that precedence constraints, congestion,
and delay propagation can change realized execution performance after a plan is
computed. Building on this execution-structure perspective, we ask a hardware-transfer
question: whether primitive motion burden and interaction-aware plan structure,
measured before execution, provide held-out predictive value for realized
wall-clock execution time beyond SoC.

\section{Study Design, Scenario Library, and Plan Corpus}\label{sec:study-design}

\subsection{Platform and task scope}
The study measures realized execution time on a fixed $7\times 7$ workcell
with seven differential-drive Pololu 3Pi+ robots. The target is not nominal planner
quality but total wall-clock completion time under the execution system
described in Section~\ref{sec:execution}. By holding the platform, executor,
and task geometry fixed, the analysis isolates which plan-side descriptors
transfer to hardware on this system.

\subsection{Scenario library}
The scenario library contains 15 scenarios, five from each of three declared
families: Empty, Medium-Random, and Bottleneck.
The Empty family acts as an obstacle-free control with minimal structural
routing constraint. The
Medium-Random family introduces moderate routing restriction through fixed
random obstacle layouts that create corridor sharing and local detours
without collapsing all motion into a single choke point. The Bottleneck family
concentrates narrow-passage dependency pressure through a fixed contested
passage geometry. We use this library to span distinct interaction regimes
rather than to approximate a broad benchmark
suite~\cite{Stern2019, Hoenig2019, Varambally2022,Chen2024}. This family balance
is the declared corpus-design choice. Within each family, candidate
start--goal assignments are generated by seeded validity rules, with CBSH2-RTC
used as the scenario-admission feasibility oracle before final scenario
identities are frozen. The resulting 15-scenario library is a pre-specified
interaction-regime sample for this platform, fixed before hardware execution
and regression fitting.

\input{tables/tab1_protocol_summary.tex}
\subsection{Plan-corpus construction}
For each frozen scenario, we instantiate the same eight plan-generation slots.
The fixed menu contains one optimal reference plan from CBSH2-RTC, four
bounded-suboptimal plans from EECBS at multiple weight settings, and three
seed-varied plans from
LaCAM3~\cite{Li2019CBS,Li2021EECBS,Okumura2023LaCAM,Okumura2024LaCAMStar,okumura2024engineeringlacam}.
Applied uniformly to all hardware corpus whose
prediction target is plan-level mean wall-clock time rather than a single run.
The declared generator menu induces cost and structural variation under a
feasible hardware budget while keeping the construction auditable: every
scenario receives the same treatment. Accordingly, the study asks whether
progressively richer plan descriptors explain realized execution time better
than SoC alone, not which planner should be declared best.

\section{Execution System, Feature Ladder, and Held-Out Evaluation}\label{sec:execution}

\subsection{Hardware execution system}
\begin{figure}[H]
  \centering
  \includegraphics[width=\columnwidth]{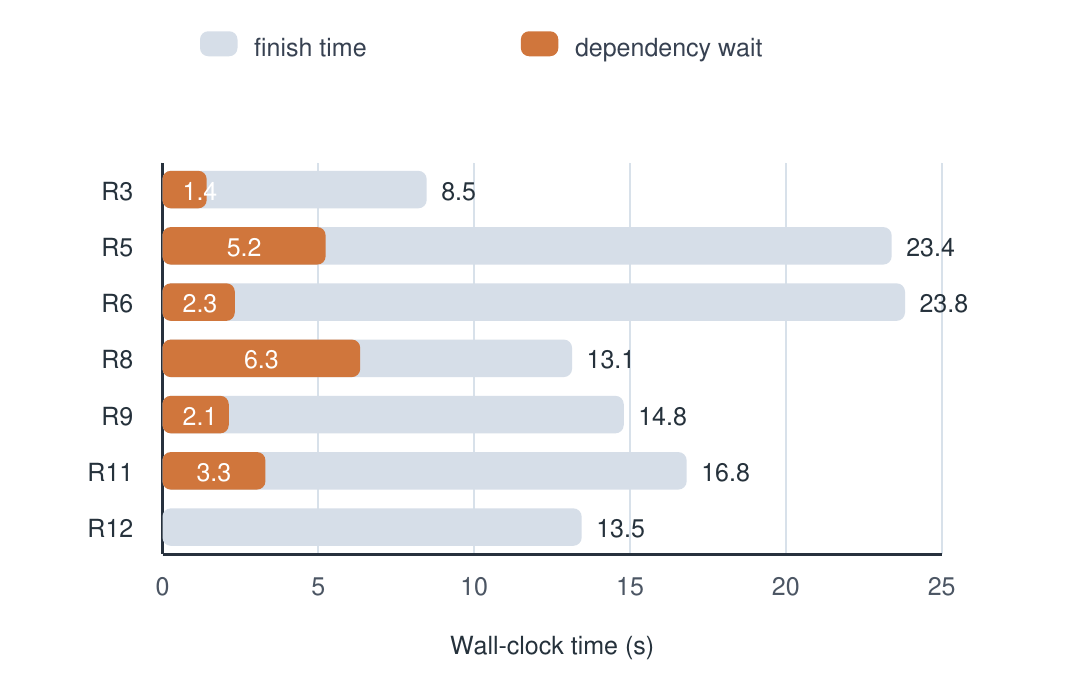}
  \caption{\textbf{Executor dependency-wait diagnostic.} In one representative
  hardware trial, dependency-gated waiting accounts for a visible share of
  several robots' wall-clock finish times. This illustrates the execution
  mechanism rather than a corpus-level result.}
  \label{fig:executor-dependency-wait}
\end{figure}

Hardware trials are executed under a precedence-faithful continuous executor
(i.e., an executor that preserves the dependency order implied by the plan, so
each stage starts only after the robot's previous stage and all required
cross-robot predecessor stages have completed). The MAPF plan is treated as a
shared logical schedule whose same-agent stage order and cross-robot
precedence constraints must be preserved during execution. However, execution
is not barrier-synchronized by a global timestep release clock. A stage may
begin only after the previous stage of that robot has completed and all
incoming cross-robot dependencies have been
satisfied~\cite{Ma2017,Hoenig2019,Su2024}. This choice keeps the execution
semantics close to the logical coordination structure of the plan while still
exposing real wall-clock delay, local contention, and stop-go behavior on
hardware, as shown in Figure~\ref{fig:executor-dependency-wait}. It also makes
the executor an explicit part of the studied system rather than a hidden
implementation detail. Accordingly, all predictive claims are
executor-conditioned: alternative release policies or switchable passing-order
Controllers may reshape realized waiting and should be evaluated as separate
execution stacks.

\subsection{Pre-execution feature extraction}
\label{sec:methods-features}

All predictors are computed before hardware execution from the MAPF plan
alone; no hardware timing outcome enters the extraction pipeline. For each
normalized plan, we first compute the nominal plan-cost and primitive-motion
features directly from the per-robot path sequences:SoC,
planner-side makespan, total turn count, consecutive-move count, and
start--stop transition count.

We then compile the same plan into an Action Dependency Graph
(ADG)~\cite{Hoenig2019} that the precedence-faithful continuous executor of
Section~\ref{sec:execution} consumes at run time. In this ADG, nodes are
per-robot stages, intra-robot edges encode the same-agent sequential order, and
inter-robot edges encode cross-robot precedence induced by shared vertex
occupancy and edge swaps under the planned ordering. The executor releases
each stage only after its intra-robot predecessor and all incoming inter-robot
predecessors have completed.

The interaction-aware features are then read directly off this ADG: the number
of inter-robot precedence edges, the number of distinct robot pairs connected
by those edges, and the depth of the longest inter-robot dependency chain.
Crowding exposure is reported separately and is not an ADG edge count; it is
computed from the co-occupancy of planned robot states over active portions of the
trajectories, and is included as a plan-trace proximity descriptor that
complements the ADG-derived edge features.

\subsection{Feature ladder}
We evaluate a predefined four-level model ladder:
\begin{itemize}
    \item \emph{Model~0 (null baseline):} a baseline with no plan-side predictive features, used only as a reference for the richer models.

    \item \emph{Model~1 (SoC only):} uses Sum of Costs (SoC, the total planned travel cost summed across all robots) as the only predictor.

    \item \emph{Model~2 (SoC + primitive motion burden):} augments SoC with four execution-aware motion features that capture how much basic motion the plan demands: planner-side makespan (the logical number of timesteps until the last robot reaches its goal), total turn count (the total number of direction changes across all robots), consecutive-move count (how often robots continue moving across successive steps without pausing), and start--stop transition count (how often robots switch between moving and waiting).

    \item \emph{Model~3 (SoC + primitive motion burden + interaction-aware coordination load):} further adds four coordination features that capture how much inter-robot interaction the plan induces: the number of cross-robot precedence edges (dependency links where one robot must wait for another), the number of distinct robot pairs they connect (how many different robot pairs are involved in such dependencies), the depth of the longest cross-robot dependency chain (the length of the longest multi-robot wait sequence), and a crowding-exposure count (how much planned motion passes through shared space at the same logical time).
\end{itemize}

Planner-side makespan is used here as a logical plan-horizon feature in timesteps rather than as a direct estimate of physical seconds; the prediction target remains realized wall-clock execution time.

\subsection{Outcome and validation}
The primary target is realized total execution wall-clock time, measured
per trial from the hardware logs. Each plan is executed four times, but
The four repeats of one plan are not treated as four independent test
rows: between-plan variance accounts for $99.4\%$ of total
execution-time variance and the median within-plan coefficient of
variation is $1.6\%$, so the plan-level mean is a stable summary and the
Repeats serve as a variability estimate rather than independent
confirmatory evidence. The confirmatory unit is therefore the plan mean.

Held-out validation is then performed at the \emph{scenario} level, not
the plan level. All plans and all repeats sharing a scenario must fall
on the same side of the train/test split, otherwise the model would see
the same start--goal layout in both training and evaluation. We use a
5-fold family-balanced scenario split: each fold holds out three
scenarios (one Empty, one Medium-Random, one Bottleneck), the ridge
ladder is fit on the remaining 12, and MAE/RMSE are computed on the
held-out plan means.

We report two trained estimators over the same M0--M3 feature ladder. The ridge
model is a ridge-regularized linear ladder fit to plan-mean outcomes, with
fold-local feature standardization, an unpenalized intercept, and ridge penalty
selection by scenario-grouped inner cross-validation inside each outer training
fold. The mixed-effects model is fit at the trial level over repeated hardware
executions and is evaluated on held-out plan means so that its MAE and RMSE are
directly comparable to the ridge model. Both estimators use the same
scenario-level family-balanced 5-fold outer holdout. The held-out unit is the
scenario, so all plans and repetitions derived from one scenario remain in the
same fold. With the 15-scenario library, each outer fold holds out three
scenarios: one Empty, one Medium-Random, and one Bottleneck.

The staged comparisons are the same for both trained models: first, whether
primitive execution burden improves on SoC alone, and second, whether
interaction-aware structure improves on the primitive-motion tier. We report
MAE and RMSE in seconds. For each stepwise comparison, scenario-blocked
Bootstrap intervals and paired scenario-level $t$ intervals summarize the
reliability of the held-out error change. Deltas are defined as higher-tier
error minus lower-tier error, so negative values indicate improvement. All
interpretations remain platform-scoped and executor-conditioned.

\section{Results}\label{sec:results}

This section reports held-out predictive performance for the same M0--M3
feature ladder under two trained models on the 480-trial hardware corpus. The
ridge model evaluates scenario-held-out plan-mean prediction. The mixed-effects
model fits repeated trial-level executions and is evaluated on held-out plan
means for the same MAE and RMSE readout. The two central comparisons are
$M_1 \rightarrow M_2$, which tests whether primitive motion burden adds signal
beyond SoC, and $M_2 \rightarrow M_3$, which tests whether interaction-aware
structure adds held-out improvement beyond the primitive-motion tier.

\input{tables/two_model_ladder_tables}

\input{tables/two_model_stepwise_reliability_tables}

Tables~\ref{tab:two-model-ladder-errors} and
\ref{tab:two-model-stepwise-reliability} present the main readout. The shared
pattern is clear: SoC alone is incomplete, and the primitive-motion tier gives
the largest reduction in held-out error. In the ridge model, $M_3$ gives a small favorable point
estimate whose intervals cross zero.

In the mixed-effects model, $M_3$ gives a
larger additional reduction, with bootstrap intervals below zero and
scenario-level $t$ intervals close to zero. Figure~\ref{fig:hardware-trial}
anchors these results in one measured hardware execution from the same corpus.

\subsection{Primitive motion burden recovers the largest shared signal}
Primitive execution burden produces the largest shared improvement beyond SoC.
In the ridge model, $M_2$ reduces MAE from $2.7626$ to $1.4196$ s (down by
$1.3430$ s, or about $48.6\%$) and RMSE from $3.4118$ to $1.9029$ s (down by
$1.5089$ s, or about $44.2\%$). The ridge $M_2-M_1$ intervals are below zero
for both metrics: MAE bootstrap 95\% CI $[-2.2096,-0.5878]$ and $t$ 95\% CI
$[-2.3043,-0.4268]$; RMSE bootstrap 95\% CI $[-2.4946,-0.5780]$ and $t$ 95\%
CI $[-2.3284,-0.4421]$, supporting the interpretation that primitive motion
burden adds reliable held-out predictive value beyond SoC. The mixed-effects
model shows the same feature-family message with even larger stepwise
reductions: MAE falls from $4.7933$ to $1.9273$ s (down by $2.8660$ s, or about
$59.8\%$) and RMSE from $6.3315$ to $2.4469$ s (down by $3.8846$ s, or about
$61.4\%$), again with both bootstrap and $t$ intervals below zero. Because
these features are computable from the offline plan before execution begins,
this result identifies primitive motion burden as the strongest missing layer
between SoC and realized hardware execution time.
\vspace{\baselineskip}

\subsection{SoC is informative but incomplete}
\begin{figure}[H]
  \centering
  \includegraphics[width=\columnwidth]{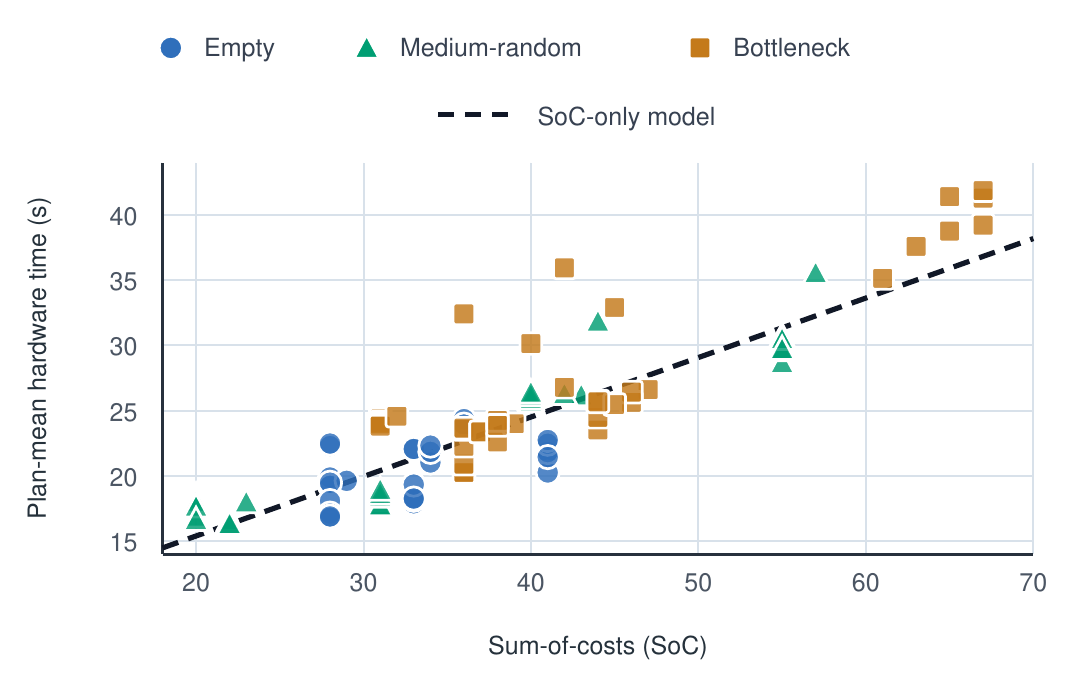}
  \caption{\textbf{SoC-only model.} Each point is a plan mean; the dashed line
  is the SoC-only model. The vertical spread shows why SoC alone is not a complete predictor of hardware time.}
  \label{fig:soc-hardware-time}
  \vspace{-8pt}
  \end{figure}
In the ridge model, SoC reduces held-out MAE from
\\
\\
\\
\\
 $4.9998$ to $2.7626$ s and
RMSE from $6.4507$ to $3.4118$ s. In the mixed-effects model, the SoC-only tier
remains close to the null baseline, with MAE $4.7933$ s and RMSE $6.3315$ s.
The combined reading is not that SoC is irrelevant. Rather, SoC is too coarse
to serve as a complete predictor for realized wall-clock cost on this platform,
especially once a repeated trial structure is modeled.

\subsection{Interaction-aware structure under the current corpus}
Adding the interaction-aware feature tier on top of the primitive-motion tier moves both estimators in the same direction. (For reference, the primitive-motion tier model~M2 adds per-plan turn count, start--stop count, and total commanded path length on top of SoC; the interaction-aware tier model~M3 adds, on top of M2, schedule-level coordination quantities such as cross-robot dependency count, dependency-chain depth, and crowding exposure.) Under ridge, MAE drops from $1.4196\,\mathrm{s}$ to $1.3969\,\mathrm{s}$ ($-0.0228\,\mathrm{s}$, ${\approx}1.6\%$) and RMSE from $1.9029\,\mathrm{s}$ to $1.7513\,\mathrm{s}$ ($-0.1516\,\mathrm{s}$, ${\approx}8.0\%$); both held-out intervals cross zero. Under the mixed-effects model, MAE drops from $1.9273\,\mathrm{s}$ to $1.4310\,\mathrm{s}$ ($-0.4963\,\mathrm{s}$, ${\approx}25.8\%$) and RMSE from $2.4469\,\mathrm{s}$ to $1.7885\,\mathrm{s}$ ($-0.6584\,\mathrm{s}$, ${\approx}26.9\%$), with bootstrap intervals below zero on both metrics, while the scenario-level $t$ intervals span MAE $[-1.0527,\,0.0493]$ and RMSE $[-1.1181,\,0.0075]$.

We therefore report the interaction-aware tier as an additional, directionally consistent layer of the model ladder rather than as a settled effect-size claim. The evidence is aligned in direction across both trained models, but the present corpus does not support a precise statement about how large this additional coordination-structure signal is on the platform. Sizing that layer cleanly is the natural target for a larger held-out scenario corpus.
\subsection{Reading the two estimators together}
The estimator comparison defines the current evidence boundary. Both ridge and mixed-effects analyses show that sum-of-costs is too coarse for realized hardware execution time and that primitive-motion burden materially improves prediction. For the interaction-aware tier, the estimates are aligned in direction but not equally resolved by the present corpus: ridge on held-out plan means does not separate the interaction-aware tier from the primitive-motion tier with intervals away from zero, whereas the mixed-effects analysis estimates a larger reduction after using the repeated executions. Given the current corpus size, we are not able to certify the reliability or precise signal magnitude of the additional interaction-aware features. The next empirical step is therefore to expand the scenarios. The joint reading is therefore that primitive motion burden is the supported missing layer between SoC and realized execution time on this platform; interaction-aware structure carries a directionally consistent additional signal whose resolution under the primary held-out comparison is bounded by the information content of the present corpus.
\section{Discussion and Limitations}\label{sec:discussion}

\noindent\textbf{How can these predictions be used in deployment, and what remains uncertain?}
Because these descriptors are computable from the offline plan before any robot moves, they can be used to score or compare candidate plans using a quantity that is closer to realized execution time than SoC alone on this hardware stack. In practice, the predicted values are best used for plan selection, or deployment-side screening among candidate plans generated for the same platform and executor, rather than as universal execution-time guarantees. The main uncertainty is scale transfer: all claims here remain platform-scoped and executor-conditioned, so transfer to larger fleets or different execution policies requires separate hardware validation.

\textbf{Limitations.}
\begin{itemize}
  \item \textit{Controlled hardware scope.} The findings are conditioned on seven robots, a $7\times7$ arena, and one precedence-faithful continuous executor, so transfer to larger fleets or different execution stacks requires new hardware evidence.
  \item \textit{Compact scenario support.} The 15 scenarios span Empty, Medium-Random, and Bottleneck regimes, but they do not constitute a broad benchmark suite.
  \item \textit{Declared corpus design.} The fixed eight-slot plan-generation menu makes the corpus auditable and repeatable, but not representative of all MAPF tasks, solvers, or deployment settings.
  \item \textit{Interpretable rather than maximal models.} The ridge and mixed-effects ladders are designed as disciplined, interpretable tests of feature families, not as upper bounds on predictive performance.
\end{itemize}

\section*{Acknowledgment}
The authors thank Prof.\ Pavankumar Tallapragada, Associate Professor at the Robert Bosch Centre for Cyber-Physical Systems (RBCCPS), Indian Institute of Science (IISc), Bengaluru, for his guidance and the resources that made this work possible. The authors also acknowledge BuildMachineLabs, a research-centric laboratory based in Bengaluru, for the initial motivation and continued support for this effort.

\section{Conclusion}\label{sec:conclusion}
This study concludes that standard MAPF plan quality is insufficient to predict
realized execution time on a physical multi-robot workcell and additional
plan-side structure is needed before execution. This is validated using a seven-robot, $7\times7$
hardware platform under a precedence-faithful continuous executor, SoC
captures an important trend but does not fully explain wall-clock completion
time. The clearest additional signal is primitive motion burden: adding
planner-side makespan, turns, consecutive moves, and start--stop transitions to
SoC gives the largest and most reliable improvement across both trained
analyses, reducing held-out MAE from $2.7626$\,s to $1.4196$\,s in the ridge
model and from $4.7933$\,s to $1.9273$\,s in the mixed-effects model.The Interaction-aware coordination structure points to a further layer of execution
signal, but the present hardware corpus does not settle its reliability or
magnitude. The interaction-aware tier improves the point estimates in both
analyses, while the strictest held-out comparison does not separate it from the
primitive-motion tier with intervals away from zero. The resulting design implication
is that execution-aware MAPF evaluation should not rely on SoC alone: even in a compact
hardware study, much of the missing execution-time signal is captured by primitive motion
descriptors before any robot moves.
\bibliographystyle{IEEEtran}
\bibliography{references}

\end{document}

%% file: figures/fig_study_overview.tex
\begin{figure}[!htbp]
\centering
\begin{tikzpicture}[
  x=1cm,
  y=1cm,
  font=\scriptsize,
  panel/.style={draw=black!65, rounded corners=1.4pt, line width=0.35pt, fill=white},
  paneltitle/.style={font=\bfseries\scriptsize, anchor=west},
  statbox/.style={draw=black!20, rounded corners=1.2pt, line width=0.3pt, fill=black!2, minimum width=1.18cm, minimum height=0.58cm, inner sep=1pt},
  gridline/.style={draw=black!22, line width=0.2pt},
  minigrid/.style={draw=black!18, line width=0.16pt}
]
\newcommand{\roboticon}[4]{%
  \begin{scope}[shift={(#1,#2)}, rotate=#3]
    \draw[line width=0.45pt, draw=black!72] (-0.22,0.15) -- (-0.22,-0.15);
    \draw[line width=0.45pt, draw=black!72] (0.22,0.15) -- (0.22,-0.15);
    \fill[#4] (0,0) circle[radius=0.22];
    \draw[line width=0.35pt, draw=black!72] (0,0) circle[radius=0.22];
    \draw[line width=0.55pt, draw=white] (0,0) -- (0.18,0);
    \fill[white] (0.20,0) circle[radius=0.035];
  \end{scope}
}
\newcommand{\drawminimap}[3]{%
  \begin{scope}[shift={(#1,#2)}, x=0.215cm, y=0.215cm]
    \fill[#3!5] (0,0) rectangle (7,7);
    \draw[line width=0.38pt, draw=black!64] (0,0) rectangle (7,7);
    \foreach \i in {1,...,6}{
      \draw[minigrid] (\i,0) -- (\i,7);
      \draw[minigrid] (0,\i) -- (7,\i);
    }
  \end{scope}
}

\path[panel, fill=blue!2] (0,4.25) rectangle (8.50,8.12);
\path[draw=black!18, fill=white, rounded corners=1.2pt, line width=0.25pt]
  (0.34,4.48) rectangle (3.50,7.64);

\begin{scope}[shift={(0.55,4.69)}, x=0.39cm, y=0.39cm]
  \fill[cyan!8] (0,0) rectangle (7,7);
  \draw[line width=0.46pt, draw=black!62] (0,0) rectangle (7,7);
  \foreach \i in {1,...,6}{
    \draw[gridline] (\i,0) -- (\i,7);
    \draw[gridline] (0,\i) -- (7,\i);
  }
  \draw[-{Latex[length=1.0mm,width=0.8mm]}, line width=0.48pt, draw=teal!75!black, rounded corners=0.4pt]
    (0.8,6.0) -- (2.2,6.0) -- (2.2,5.1);
  \draw[-{Latex[length=1.0mm,width=0.8mm]}, line width=0.48pt, draw=orange!80!black, rounded corners=0.4pt]
    (3.0,6.2) -- (3.0,4.7) -- (4.1,4.7);
  \draw[-{Latex[length=1.0mm,width=0.8mm]}, line width=0.48pt, draw=blue!70!black, rounded corners=0.4pt]
    (5.6,5.4) -- (4.7,5.4) -- (4.7,6.3);
  \draw[-{Latex[length=1.0mm,width=0.8mm]}, line width=0.48pt, draw=red!70!black, rounded corners=0.4pt]
    (1.4,3.6) -- (2.4,3.6) -- (2.4,2.7);
  \draw[-{Latex[length=1.0mm,width=0.8mm]}, line width=0.48pt, draw=olive!70!black, rounded corners=0.4pt]
    (4.1,3.4) -- (5.6,3.4) -- (5.6,2.6);
  \draw[-{Latex[length=1.0mm,width=0.8mm]}, line width=0.48pt, draw=violet!75!black, rounded corners=0.4pt]
    (6.1,2.0) -- (5.0,2.0) -- (5.0,1.0);
  \draw[-{Latex[length=1.0mm,width=0.8mm]}, line width=0.48pt, draw=magenta!65!black, rounded corners=0.4pt]
    (2.8,1.1) -- (1.7,1.1) -- (1.7,2.0);
  \roboticon{0.8}{6.0}{20}{teal!70!black}
  \roboticon{3.0}{6.2}{70}{orange!80!black}
  \roboticon{5.6}{5.4}{-10}{blue!70!black}
  \roboticon{1.4}{3.6}{155}{red!70!black}
  \roboticon{4.1}{3.4}{45}{olive!70!black}
  \roboticon{6.1}{2.0}{225}{violet!75!black}
  \roboticon{2.8}{1.1}{300}{magenta!65!black}
\end{scope}
\node[paneltitle] at (0.28,7.84) {A. Fixed 7x7 hardware workcell};

\node[anchor=north west, font=\bfseries\scriptsize, text=black!84, inner sep=0pt] at (3.85,7.22) {Hardware scope for the plan corpus};
\node[anchor=north west, align=left, text width=4.18cm, font=\scriptsize, text=black!70, inner sep=0pt] at (3.85,6.83)
  {All generated MAPF plans are executed on the same physical arena and seven-robot fleet.};
\node[statbox, fill=blue!7, draw=blue!42!black, minimum width=1.38cm] at (4.48,5.08) {};
\node[font=\bfseries\scriptsize, text=blue!62!black] at (4.48,5.20) {$7\times7$};
\node[font=\tiny] at (4.48,4.95) {workcell};
\node[statbox, fill=orange!9, draw=orange!55!black, minimum width=2.92cm] at (6.75,5.08) {};
\node[font=\bfseries\scriptsize, text=orange!70!black] at (6.75,5.20) {7 differential-drive};
\node[font=\tiny] at (6.75,4.95) {robots};

\path[panel, fill=black!1] (0,1.72) rectangle (8.50,4.05);
\node[paneltitle] at (0.18,3.79) {B. Scenario families};

\drawminimap{0.55}{2.08}{teal}
\drawminimap{3.50}{2.08}{orange}
\drawminimap{6.45}{2.08}{red}

\begin{scope}[shift={(3.50,2.08)}, x=0.215cm, y=0.215cm]
  \foreach \row/\col in {0/1,2/4,4/1,5/2,5/5,6/0}{
    \fill[black!72] (\col,{6-\row}) rectangle ++(1,1);
  }
\end{scope}
\begin{scope}[shift={(6.45,2.08)}, x=0.215cm, y=0.215cm]
  \foreach \row/\col in {0/3,1/3,2/3,4/3,5/3,6/3}{
    \fill[black!72] (\col,{6-\row}) rectangle ++(1,1);
  }
  \fill[green!18] (3,3) rectangle ++(1,1);
\end{scope}

\draw[line width=0.65pt, draw=teal!60!black] (0.95,2.00) -- (1.65,2.00);
\draw[line width=0.65pt, draw=orange!70!black] (3.79,2.00) -- (4.71,2.00);
\draw[line width=0.65pt, draw=red!60!black] (6.80,2.00) -- (7.60,2.00);
\node[font=\bfseries\tiny, text=black!86] at (1.30,1.84) {Empty};
\node[font=\bfseries\tiny, text=black!86] at (4.25,1.84) {Medium-random};
\node[font=\bfseries\tiny, text=black!86] at (7.20,1.84) {Bottleneck};
\end{tikzpicture}
\vspace{-0.35em}
\caption[Hardware platform and scenario-family overview.]{\textbf{Hardware
platform and scenario-family overview.}}
\label{fig:overview}
\end{figure}

%% file: tables/tab1_protocol_summary.tex
\begin{table}[H]
\centering
\caption{Study protocol and hardware corpus.}
\label{tab:protocol-summary}
\begin{tabular}{@{}>{\raggedright\arraybackslash}p{0.34\columnwidth}>{\raggedright\arraybackslash}p{0.58\columnwidth}@{}}
\toprule
Element & Value \\
\midrule
Platform & Seven differential-drive robots on a $7\times 7$ physical workcell \\
Scenario library & 15 scenarios: 5 Empty, 5 Medium-Random, 5 Bottleneck \\
Plan-generation menu & 1 CBSH2-RTC; 4 EECBS weights; 3 LaCAM seeds per scenario \\
Corpus construction & Seeded scenario freeze; fixed menu for every scenario \\
Plans per scenario & 8 planned MAPF solutions \\
Plan corpus & 120 planned MAPF plans \\
Hardware corpus & 480 scheduled executions; four per plan \\
Hardware repetitions & 4 scheduled executions per plan \\
Evaluation target & Plan-level mean execution time \\
Execution monitor & Precedence-faithful continuous executor \\
Measured target & Total execution wall-clock time \\
Held-out evaluation & Scenario-level family-balanced 5-fold split \\
Estimators & Ridge plan-mean ladder and trial-level mixed-effects ladder \\
Metrics & Held-out MAE and RMSE in seconds \\
\bottomrule
\end{tabular}
\end{table}

%% file: tables/two_model_ladder_tables.tex
\begin{table*}[t]
\centering
\caption{Held-out model-ladder errors for the ridge and mixed-effects models
on the hardware corpus. Both panels use the same M0--M3 feature
ladder. Delta columns are relative to the previous model tier; negative values
indicate better prediction.}
\label{tab:two-model-ladder-errors}
\renewcommand{\arraystretch}{1.12}
\scriptsize
\begin{tabular*}{\textwidth}{@{\extracolsep{\fill}}llrrrrl@{}}
\toprule
\multicolumn{7}{@{}l}{\textbf{A. Held-out ridge error by model tier}}\\
\midrule
Model & Feature set & MAE & $\Delta$MAE & RMSE & $\Delta$RMSE & Interpretation \\
\midrule
$M_0$ & Null & $4.9998$ & -- & $6.4507$ & -- & baseline \\
$M_1$ & SoC only & $2.7626$ & $-2.2372$ & $3.4118$ & $-3.0389$ & informative but incomplete \\
$M_2$ & SoC + primitive motion burden & $1.4196$ & $-1.3430$ & $1.9029$ & $-1.5089$ & large improvement \\
$M_3$ & SoC + motion + interaction-aware structure & $1.3969$ & $-0.0228$ & $1.7513$ & $-0.1516$ & small point gain \\
\midrule
\multicolumn{7}{@{}l}{\textbf{B. Held-out mixed-effects error by model tier}}\\
\midrule
Model & Feature set & MAE & $\Delta$MAE & RMSE & $\Delta$RMSE & Interpretation \\
\midrule
$M_0$ & Null & $5.0352$ & -- & $6.4750$ & -- & baseline \\
$M_1$ & SoC only & $4.7933$ & $-0.2419$ & $6.3315$ & $-0.1434$ & informative but incomplete \\
$M_2$ & SoC + primitive motion burden & $1.9273$ & $-2.8660$ & $2.4469$ & $-3.8846$ & large improvement \\
$M_3$ & SoC + motion + interaction-aware structure & $1.4310$ & $-0.4963$ & $1.7885$ & $-0.6584$ & additional reduction \\
\bottomrule
\end{tabular*}
\end{table*}

%% file: tables/two_model_stepwise_reliability_tables.tex
\begin{table*}[t]
\centering
\caption{Stepwise held-out error changes for the ridge and mixed-effects models
on the hardware corpus. Negative deltas mean the richer model has
better prediction. Bootstrap intervals resample scenario blocks; $t$ intervals
use one paired delta per scenario.}
\label{tab:two-model-stepwise-reliability}
\renewcommand{\arraystretch}{1.12}
\scriptsize
\begin{tabular*}{\textwidth}{@{\extracolsep{\fill}}llrllll@{}}
\toprule
\multicolumn{6}{@{}l}{\textbf{A. Tested ridge increments with two interval estimates}}\\
\midrule
Comparison & Metric & $\Delta$ error & Bootstrap 95\% CI & $t$ 95\% CI & Interpretation \\
\midrule
$M_2$ vs. $M_1$ & MAE  & $-1.3430$ & $[-2.2096,-0.5878]$ & $[-2.3043,-0.4268]$ & supported \\
$M_2$ vs. $M_1$ & RMSE & $-1.5089$ & $[-2.4946,-0.5780]$ & $[-2.3284,-0.4421]$ & supported \\
$M_3$ vs. $M_2$ & MAE  & $-0.0228$ & $[-0.4995,0.4740]$  & $[-0.5751,0.5188]$  & not confirmed \\
$M_3$ vs. $M_2$ & RMSE & $-0.1516$ & $[-0.6296,0.4421]$  & $[-0.6756,0.4298]$  & not confirmed \\
\midrule
\multicolumn{6}{@{}l}{\textbf{B. Tested mixed-effects increments with two interval estimates}}\\
\midrule
Comparison & Metric & $\Delta$ error & Bootstrap 95\% CI & $t$ 95\% CI & Interpretation \\
\midrule
$M_2$ vs. $M_1$ & MAE  & $-2.8660$ & $[-4.6168,-1.4106]$ & $[-4.8200,-1.0468]$ & supported \\
$M_2$ vs. $M_1$ & RMSE & $-3.8846$ & $[-5.5813,-1.8878]$ & $[-4.6962,-1.0334]$ & supported \\
$M_3$ vs. $M_2$ & MAE  & $-0.4963$ & $[-1.0122,-0.0385]$ & $[-1.0527,0.0493]$  & bootstrap below zero; $t$ near zero \\
$M_3$ vs. $M_2$ & RMSE & $-0.6584$ & $[-1.0969,-0.1936]$ & $[-1.1181,0.0075]$  & bootstrap below zero; $t$ near zero \\
\bottomrule
\end{tabular*}
\end{table*}